\documentclass[runningheads]{llncs}
\usepackage{amssymb}
\usepackage{graphicx}
\usepackage{caption}
\usepackage{subcaption}
\usepackage[utf8]{inputenc}
\usepackage{algorithm}
\usepackage{algpseudocode}
\usepackage[T1]{fontenc}
\usepackage{amsmath}
\usepackage{multirow} 
\usepackage{comment}
\usepackage{pdflscape} 
\usepackage{array}     
\usepackage{authblk}
\usepackage{dsfont}
\usepackage{lscape} 
\usepackage{longtable} 
\title{Evaluating Model Robustness Using Adaptive Sparse L0 Regularization}

\author{Weiyou Liu\inst{1}\thanks{Email: weiyou.liu@adelaide.edu.au} \and
        Zhengyang Li\inst{1}\thanks{Email: zhengyang.li01@adelaide.edu.au} \and
        Weitong Chen\inst{1}\thanks{Email: weitong.chen@adelaide.edu.au}}

\institute{University of Adelaide, Australia}
\begin{document}

\maketitle

\begin{abstract}
Deep Neural Networks (DNNs) have demonstrated remarkable success in various domains but remain susceptible to adversarial examples: slightly altered inputs designed to induce misclassification. While adversarial attacks typically optimize under \(L_p\)-norm constraints, attacks based on the \(L_0\)-norm, which prioritize input sparsity, are less studied due to their complex, non-convex nature. These sparse adversarial examples challenge existing defenses by altering a minimal subset of features, potentially uncovering more subtle DNN weaknesses. However, the current \(L_0\)-norm attack methodologies face a trade-off between accuracy and efficiency—either precise but computationally intense or expedient but imprecise. This paper proposes a novel, scalable, and effective approach to generate adversarial examples of the \(L_0\) norm, aimed at refining the robustness evaluation of DNNs against such perturbations.

\end{abstract}

\section{Introduction}

Deep Neural Networks (DNNs) have achieved unprecedented success in a myriad of applications, ranging from image recognition to natural language processing. However, their susceptibility to adversarial attacks, where subtly modified inputs can lead to erroneous model outputs, poses significant challenges to their reliability and security\cite{szegedy2014,goodfellow2015}. These adversarial perturbations not only compromise the reliability of DNNs in real-world applications but also expose underlying vulnerabilities in their decision-making mechanisms\cite{carlini2017}. For example, in autonomous driving systems, adversarial attacks can manipulate sensor inputs, potentially leading to catastrophic outcomes.

Traditionally, the generation of adversarial examples has focused on perturbations constrained by \(L_2\) norms, which primarily consider the magnitude of changes\cite{biggio2013,chen2018}. However, recent studies have shifted the focus towards \(L_0\) and \(L_\infty\) norms, led by insights from Croce \& Hein\cite{croce2021a,croce2021b}, which highlight the efficacy of sparse perturbations. These perturbations are less perceptible, yet potent, challenging the conventional understanding and prompting a reevaluation of attack strategies emphasising sparsity.

Applying \(L_0\) norm attacks, which aim to alter the minimal number of input features, introduces complex challenges due to their non-convex and non-differentiable nature\cite{osborne2000b}. This has led to the development of innovative optimisation strategies that balance sparsity with attack efficacy. Techniques by Addepalli et al. \cite{addepalli2022} and Bach et al. \cite{bach2012} have significantly advanced our understanding of these dynamics, illustrating the delicate interplay between sparsity and robustness. For example, in the field of medical diagnostics, sparse adversarial perturbations can alter critical diagnostic features, leading to misdiagnosis while remaining undetected.

However, current \(L_0\) norm attack methodologies face a significant trade-off between accuracy and efficiency. Many existing solutions either achieve high accuracy but at the cost of computational intensity or are computationally efficient but lack precision in generating effective adversarial examples. This trade-off limits their practical applicability, especially in real-time or resource-constrained environments. For instance, high computational demands can render an attack infeasible in real-time applications like fraud detection in financial transactions.

We propose a novel Adaptive Sparse and Lightweight Optimisation (ASLO) approach to address these challenges. Our method dynamically adjusts perturbation parameters in real time based on model feedback, optimising for both sparsity and attack efficacy. This real-time adaptivity allows ASLO to maintain high attack success rates while minimising perturbations, achieving a better balance between accuracy and efficiency compared to existing methods. By adjusting perturbations dynamically, ASLO can efficiently deceive models in a variety of applications, from cybersecurity to speech recognition systems.

The major contributions of this paper are as follows:
\begin{itemize}
    \item We introduce the ASLO strategy, which optimizes \(L_0\)-norm perturbations with real-time adaptivity.
    \item We demonstrate that ASLO achieves a superior balance between attack effectiveness and computational efficiency.
    \item We validate the robustness and generalizability of ASLO across multiple datasets and model architectures.
\end{itemize}

Through these contributions, our aim is to provide a more robust framework for evaluating and enhancing the adversarial robustness of DNNs, addressing the critical need for secure and reliable AI systems in various high-stakes applications.

\section{Related Work}

\subsection{Adversarial Attacks on Time Series Classification}

The field of Time Series Classification (TSC) has seen significant advancements with the integration of deep learning techniques, such as Convolutional Neural Networks (CNNs) and Recurrent Neural Networks (RNNs). These advancements have enabled models to effectively capture complex temporal dependencies and patterns in time series data. However, the increasing deployment of TSC models in critical sectors, such as healthcare and finance, has highlighted their vulnerability to adversarial attacks. These attacks, designed to subtly manipulate time series data, pose serious security and reliability challenges. Understanding the specific threats to TSC models is crucial for developing robust defense mechanisms.

\subsection{Sparse Adversarial Perturbations}

In the context of adversarial attacks, sparsity—governed primarily by the \(L_0\) norm—emerges as a crucial yet underexplored dimension. Sparse adversarial examples, by altering a minimal number of input features, present a stealthier threat compared to their non-sparse counterparts. The strategic advantage of sparsity in evading detection and enhancing the transferability of attacks has been highlighted in studies by Papernot et al.\cite{Papernot2016} and Tramèr and Boneh\cite{Tramer2019}. These insights are essential for developing attacks that are both effective and difficult to detect, especially in time series data where perturbations must maintain temporal coherence.

\subsection{Challenges in \(L_0\) Norm Optimization}

Applying \(L_0\) norm attacks introduces complex challenges due to their non-convex and non-differentiable nature\cite{osborne2000b}. This has driven the development of innovative optimisation strategies that balance sparsity with attack effectiveness. For example, techniques by Addepalli et al.\cite{addepalli2022} and Bach et al.\cite{bach2012} illustrate the delicate interplay between sparsity and robustness. These approaches are critical for TSC models, where maintaining the integrity of the time series structure while introducing minimal changes is essential.

\subsection{Existing Approaches and Limitations}

Existing \(L_0\) norm attack methodologies often face a trade-off between accuracy and efficiency. Many methods achieve high accuracy but at the cost of computational intensity, or are computationally efficient but lack precision in generating effective adversarial examples. This trade-off limits their practical applicability, especially in real-time or resource-constrained environments. For instance, Louizos et al.\cite{Louizos2018} proposed differentiable approximations of the \(L_0\) norm to facilitate the development of sparser neural network architectures, while Schmidt et al.\cite{Schmidt2018} explored \(L_0\) regularisation for adversarial robustness, highlighting the intersection of sparsity and security.

\section{Methodology}

\subsection{Overview of ASLO Strategy}
The Adaptive Sparse and Lightweight Optimization (ASLO) strategy aims to generate adversarial examples by optimizing the \(L_0\) norm, focusing on achieving sparsity. The core idea is to dynamically adjust perturbation parameters in real-time based on model feedback. This real-time adaptivity allows the ASLO method to create perturbations that are both minimal and effective, maintaining the natural appearance of the data while misleading the model.

To illustrate, consider a scenario where an attacker wants to subtly alter a time series data point to deceive a classification model. Traditional methods might significantly change many data points, making the attack detectable. However, ASLO adjusts its perturbation strategy based on how the model responds to initial changes, ensuring that only a few points are altered in a way that maximises the chance of misclassification.

\subsection{Basic Principles of ASLO}
The ASLO method is designed to balance the trade-off between perturbation sparsity and attack efficacy. Traditional \(L_0\) norm attacks aim to modify the fewest possible input features, but this approach is often non-differentiable and challenging to optimize. ASLO introduces a differentiable approximation of the \(L_0\) norm, enabling the use of gradient-based optimization techniques.

\subsection{Mechanism of ASLO}
The ASLO method employs a differentiable approximation of the \(L_0\) norm, which enables the use of gradient descent for optimisation. The approximation is defined as follows:
\[
\hat{l}_0(\delta, \sigma) = \sum_{i=1}^{d} \frac{\delta_i^2}{\delta_i^2 + \sigma^2}
\]
Here, \(\delta_i\) represents the perturbation applied to feature \(i\), and \(\sigma\) controls the sparsity of the perturbation. The term \(\frac{\delta_i^2}{\delta_i^2 + \sigma^2}\) acts as a smooth approximation to the indicator function that would be 1 if \(\delta_i \neq 0\) and 0 if \(\delta_i = 0\). This smoothness allows for gradient-based optimisation, which is typically not possible with the true \(L_0\) norm due to its non-differentiability. By adjusting \(\sigma\), the degree of sparsity can be controlled. A smaller \(\sigma\) results in a higher penalty for non-zero \(\delta_i\), promoting sparsity.

\subsection{Algorithmic Implementation}
The implementation of ASLO involves iteratively adjusting \(\sigma\) and optimizing the perturbations. The steps are as follows:

\begin{algorithm}
\caption{Optimized Adaptive Adjustment of \( \sigma \) in ASLO}
\begin{algorithmic}[1]
\State Initialize \( \sigma^{(0)} \); set decay rate \( \eta_d < 1 \) and increase rate \( \eta_i > 1 \)
\State Prepare the validation set and initialize the model \( f \) with its initial performance metric
\For{\( t = 0, 1, 2, \ldots \) until convergence}
    \State Generate perturbations \( \delta^{(t)} \) for samples using the current \( \sigma^{(t)} \)
    \State Evaluate model performance and compute the objective function \( J(\delta^{(t)}; x, y, \theta, \lambda, \sigma^{(t)}) \)
    \If{model performance improves and \( J < J^* \)}
        \State Reduce \( \sigma \) to increase sparsity: \( \sigma^{(t+1)} = \eta_d \cdot \sigma^{(t)} \)
    \Else
        \State Increase \( \sigma \) to explore more effective perturbations: \( \sigma^{(t+1)} = \eta_i \cdot \sigma^{(t)} \)
    \EndIf
    \State Optimize \( J \) using gradient descent
\EndFor
\State Return optimized perturbations
\end{algorithmic}
\end{algorithm}

The objective function \( J \) integrates the predictive accuracy of the model with the sparsity of the perturbations. It is defined as:
\[
J(\delta; x, y, \theta, \lambda, \sigma) = L(f(x+\delta, \theta), y) + \lambda \hat{l}_0(\delta, \sigma)
\]
where \( L \) measures the misclassification error, and \( \lambda \hat{l}_0 \) penalises non-sparse perturbations. Specifically, \( L(f(x+\delta, \theta), y) \) represents the loss function that measures how well the perturbed input \( x+\delta \) fools the model \( f \). The term \( \lambda \hat{l}_0(\delta, \sigma) \) adds a penalty for non-sparse perturbations, with \(\lambda\) controlling the trade-off between attack efficacy and perturbation sparsity.

\subsection{Example and Detailed Explanation}
To better understand how ASLO works, consider the following example:

Assume we have a time series input \( x \) and a model \( f \). Our goal is to generate a perturbation \( \delta \) that misleads the model while altering as few data points as possible.

\begin{enumerate}
    \item \textbf{Initialization}: We start with an initial \(\sigma^{(0)}\) and set decay and increase rates \(\eta_d\) and \(\eta_i\). These parameters control how we adjust the sparsity of the perturbations during the optimization process.

    \item \textbf{Perturbation Generation}: For each iteration \( t \), we generate a perturbation \(\delta^{(t)}\) for the input samples using the current value of \(\sigma^{(t)}\). The perturbation is calculated to maximize the likelihood of misclassification while maintaining minimal changes to the original input.

    \item \textbf{Evaluation and Adjustment}: We evaluate the model's performance with the perturbed input \( x + \delta^{(t)} \) and compute the objective function \( J(\delta^{(t)}; x, y, \theta, \lambda, \sigma^{(t)}) \). If the perturbation improves model performance (i.e., increases the likelihood of misclassification) and the objective function \( J \) is less than a predefined threshold \( J^* \), we reduce \(\sigma\) to increase the sparsity of the perturbation: 
    \[
    \sigma^{(t+1)} = \eta_d \cdot \sigma^{(t)}
    \]
    Otherwise, we increase \(\sigma\) to explore more effective perturbations:
    \[
    \sigma^{(t+1)} = \eta_i \cdot \sigma^{(t)}
    \]

    \item \textbf{Optimization}: Using gradient descent, we optimize the objective function \( J \) to find the perturbation \( \delta \) that effectively misleads the model while ensuring sparsity. The objective function combines the model's predictive accuracy and the sparsity of the perturbations:
    \[
    J(\delta; x, y, \theta, \lambda, \sigma) = L(f(x+\delta, \theta), y) + \lambda \hat{l}_0(\delta, \sigma)
    \]
    Here, \( L(f(x+\delta, \theta), y) \) represents the misclassification error, and \( \lambda \hat{l}_0(\delta, \sigma) \) is the penalty for non-sparse perturbations.

    \item \textbf{Iterative Process}: We repeat the above steps until convergence, i.e., until the perturbations become stable and the objective function \( J \) cannot be further minimized.
\end{enumerate}

For instance, consider a time series input \( x = [1, 2, 3, 4] \) and an initial perturbation \(\delta = [0.1, -0.2, 0.3, -0.4]\). Over several iterations, we adjust \(\delta\) to find the optimal sparse perturbation that maximises misclassification while minimising the changes to the original input. By iteratively updating \(\sigma\), the ASLO method ensures that the final perturbation is both effective and minimally invasive, thus achieving the desired balance between attack efficacy and perturbation sparsity.

\section{Experiment}

This section delineates our experimental setup, which is bifurcated into two principal parts. Initially, we scrutinize the efficacy of the Adaptive Sparse (AS) regularization method under controlled conditions. Subsequently, we extend the application of the AS regularization method to prevalent adversarial attack techniques, facilitating a comparative analysis of its effectiveness.

\subsection{Dataset}

Our investigation leverages the UCR Archive-2018 dataset, encompassing 128 distinct time series datasets across diverse sectors, including healthcare, agriculture, finance, and engineering. This compilation served as a solid foundation for our analysis, with each dataset partitioned into training and testing subsets to ensure a comprehensive evaluation.

\subsection{Part 1: Evaluation of the Adaptive Sparse Regularization Method}

The initial phase of our experimental study focuses on evaluating the Adaptive Sparse L0 (ASL0) regularization method. This novel approach aims to enhance model robustness by incorporating sparsity into adversarial perturbations, potentially reducing their detectability while maintaining or improving the attack's effectiveness.

\subsubsection{Experimental Setup}

To rigorously evaluate the ASL0 regularization method, we employ a controlled variable approach that ensures uniformity in the adversarial attack framework while varying the regularization technique. This approach allows for a direct comparison of the impact of ASL0 against traditional regularization methods, such as L1 and L2 regularization.

The experimental setup involves three key components:

\begin{enumerate}
    \item \textbf{Adaptive Sparse L0 Regularization (ASL0)}: The ASL0 method dynamically adjusts the sparsity level of adversarial perturbations based on real-time feedback from the model. This method's adaptability is evaluated across different settings to assess its effectiveness in generating minimal yet impactful perturbations that can deceive the target model.
    \item \textbf{Gradient-Based Iterative Perturbation Generation}: Utilizing the Carlini \& Wagner (CW) loss function, known for its precision in creating adversarial examples with minimal deviation from the original input, ensures the generation of subtle and effective adversarial examples. The integration of the CW method with various regularization approaches, including ASL0, allows for a comprehensive investigation into the balance between perturbation stealth and effectiveness.
    \item \textbf{Uniform Regularization Coefficient}: A uniform regularization coefficient of 0.00001 is applied across all experimental trials. This coefficient, selected based on preliminary experiments, strikes an optimal balance between perturbation magnitude and imperceptibility. Standardizing the regularization coefficient ensures a fair comparison among different regularization techniques, accurately assessing their impact on the sparsity, detectability, and overall effectiveness of adversarial examples.
\end{enumerate}

\subsubsection{Theoretical Rationale}

Our controlled variable approach enhances experimental rigor by isolating the regularization technique while keeping other factors constant, allowing us to directly attribute differences in model performance and adversarial robustness to the regularization method. The use of the CW loss function ensures high-quality adversarial examples that minimally deviate from original inputs, facilitating a fair evaluation of regularization efficacy. Additionally, applying a uniform regularization coefficient across all trials mitigates hyperparameter variability, providing a consistent basis for comparison and enhancing the reliability of our findings.

\subsubsection{Objective}

The primary objective is to systematically evaluate the adaptive \(L_0\) approximation method's effectiveness in generating adversarial examples that are difficult to detect yet highly effective in causing misclassification. Our controlled and comparative approach aims to highlight ASL0's advantages over other regularization techniques, offering valuable insights into how these methods can enhance the stealth and efficiency of adversarial attacks, thereby contributing to the broader discussion on model robustness and security against adversarial threats.

\subsection{Part 2: Application of ASL0 Regularization Across Common Adversarial Attack Methods}

Given the iterative nature of our innovative ASL0 regularization approach, our experiments were specifically tailored to include adversarial attack methods that generate perturbations through iterative processes. This focus ensures a direct relevance to the underlying mechanism of ASL0 regularization, enabling a nuanced examination of its impact across various adversarial frameworks.

\subsubsection{ASL0 Regularization Explained}

The ASL0 method introduces an adaptive regularisation approach that dynamically adjusts the sparsity of perturbations based on real-time feedback from the model during the iterative process. By incorporating ASL0 regularisation, perturbations are constrained not only by their magnitude but also by their sparsity. This dual constraint ensures that the generated adversarial examples are both subtle and effective. The ASL0 method's potential lies in its versatility, as it can be applied to any gradient-based iterative adversarial attack method, enhancing their ability to produce stealthy perturbations.

\subsubsection{Baseline Methods}

To comprehensively evaluate the effectiveness of ASL0 regularization, we applied it across several baseline adversarial attack methods:

\begin{itemize}
    \item \textbf{GM\_PGD (Gradient Method with Projected Gradient Descent)}: Generates perturbations in alignment with the gradient's direction, confined within a specified range to ensure stealth and efficacy.
    \item \textbf{GM\_PGD(Gradient Method with Projected Gradient Descent)\_L2}: A variation of GM\_PGD, constraining perturbation magnitudes with the \(L_2\) norm to minimize sample alteration while deceiving the model.
    \item \textbf{CW (Carlini \& Wagner) Attack}: An attack method that optimizes perturbations using a specific loss function to find the minimal perturbation necessary to mislead the model.
    \item \textbf{CW (Carlini \& Wagner)\_L2 Attack}: A variation of the CW attack, constraining perturbation magnitudes with the \(L_2\) norm to maintain sample integrity while misleading the model.
\end{itemize}

\subsubsection{Experimental Setup}

To assess the impact of ASL0 regularization on these baseline methods, we performed a series of experiments involving multiple models:

\begin{itemize}
    \item \textbf{Models Tested}: Inception, MACNN, Resnet18, LSTMFC.
    \item \textbf{Evaluation Metrics}: Attack Success Rate (ASR), Mean Success Distance, Overall Mean Distance, and Close to Zero Count.
    \item \textbf{Process}: For each model and baseline method, we generated adversarial examples with and without ASL0 regularization. We then evaluated the performance using the aforementioned metrics to determine the impact of ASL0 regularization on the effectiveness and stealth of the adversarial examples.
\end{itemize}

\subsubsection{Evaluation Metrics Overview}

To provide a comprehensive evaluation of the adversarial attacks, we employed the following metrics:

\paragraph{Attack Success Rate (ASR):}
ASR assesses the effectiveness of adversarial attacks in misleading the model, defined by the ratio of successful attacks to total attempts:
\begin{equation}
ASR = \frac{Count\_Success}{Count\_Success + Count\_Fail}
\end{equation}

\paragraph{Distance Metrics:}
To quantify the extent of perturbations, we consider the mean distance between original and perturbed samples for successful attacks:
\begin{equation}
MeanSuccessDistance = \frac{1}{Count\_Success} \sum \left\| X_{success}' - X_{success} \right\|_2
\end{equation}

\paragraph{Perturbation Sparsity:}
Sparsity evaluates the effectiveness of \(L_0\) approximation by counting perturbation values nearing zero, based on a predefined threshold (\(\epsilon\)):
\begin{equation}
CloseToZeroCount = \sum_{i=1}^{n} \mathds{1}(|r_i| < \epsilon)
\end{equation}
where \(\mathbb{1}\) is the indicator function, \(r_i\) the perturbation value, \(n\) the total number of values, and \(\epsilon\) is a small threshold (e.g., \(1e-6\)), indicating values approaching zero.

\paragraph{Threshold Definition for Sparsity:}
The threshold (\(\epsilon\)) for assessing sparsity is set to a minimal value to identify perturbation values close to zero:
\begin{equation}
\epsilon = 1e-6, \quad CloseToZeroCount = \sum \mathbb{1}(\left| r \right| < \epsilon)
\end{equation}
Given that the ASL0 method optimizes perturbation sparsity using approximate values, a minimal threshold is used instead of an exact zero value to more accurately reflect the near-zero nature of the optimized perturbations.

This selection of metrics offers a detailed evaluation of adversarial attacks, from success rates and perturbation extents to the precision of sparsity achieved through \(L_0\) approximation, thoroughly depicting the effectiveness and nuances of the attack strategy.

\section{Experimental Analysis}

\subsection{Part 1: Evaluation of the Adaptive Sparse Regularization Method}

\subsubsection{Key Observations and Discussion}

In this section, we comprehensively evaluate the Adaptive Sparse L0 (ASL0) regularization method. The focus is on its ability to produce significantly sparse perturbations, ensure a high Adversarial Success Rate (ASR), and minimize the perturbation distance required for successful attacks. We compare ASL0 with traditional regularization methods, such as L1 and L2, emphasizing computational efficiency and ASL0's dynamic adjustment mechanism.

Our analysis of the dataset reveals several critical insights into the performance of ASL0 compared to L1 and L2 methods. Although ASL0 occasionally demonstrates slightly less sparsity than L1 regularization, its adaptive feature, which dynamically adjusts sparsity based on the attack's effectiveness, provides several significant advantages:

\begin{itemize}
    \item \textbf{Adaptive Sparsity Adjustment:} ASL0 regularization incorporates an innovative mechanism that dynamically modulates the sparsity level in response to the attack's success. This ensures optimal deception with minimal perturbations on the target model.
    \item \textbf{Adversarial Success Rate (ASR):} Regularization typically reduces the attack success rate slightly, favoring increased stealth. However, ASL0 either maintains or improves the ASR compared to L1 regularization in most instances.
    \item \textbf{Distance for Successful Attacks:} By prioritizing sparsity, ASL0 significantly lowers the perturbation distance needed for successful model deception. This balance between minimal perturbation and high misclassification effectiveness represents an optimal compromise.
    \item \textbf{Computational Efficiency:} The notable sparsity of perturbations generated by ASL0 enhances computational efficiency. Requiring fewer computational resources for perturbation processing, ASL0 presents an economically viable option for producing adversarial examples.
\end{itemize}

Our experimental analysis demonstrates significant improvements with the ASL0 regularization method. ASL0 enhances attack stealth by increasing perturbation sparsity and reducing the distance required for successful attacks without significantly compromising the success rate. Its adaptive mechanism effectively balances sparsity, attack effectiveness, and computational demands, making it a valuable tool for identifying vulnerabilities in machine learning models.

As shown in Table \ref{tbl:model_performance}, ASL0 maintains the Adversarial Success Rate (ASR) while significantly reducing the perturbation distance required for successful attacks, thus enhancing stealth. This balance between minimal perturbation, high misclassification effectiveness, and computational efficiency highlights ASL0's technical capabilities and suggests new avenues for the security analysis of machine learning models.

\subsection{Part 2: Application of AS Regularization Across Common Adversarial Attack Methods}

Testing the application of AS regularization across various adversarial attack methods, including GM\_PGD, GM\_PGD\_L2, CW, CW\_L2, reveals several insights. The use of AS\_GM\_PGD and AS\_CW methods does not significantly impact the Attack Success Rate (ASR) compared to their respective non-AS regularized counterparts (GM\_PGD and CW). However, it considerably reduces the required success distance, as depicted in Figure~\ref{fig:as_gmpgd_advantages}. The upper graph illustrates the disparities in ASR between the attack methods, and the lower graph depicts the disparities in the success distance of the perturbations. This phenomenon suggests that the AS method enhances the sparsity of perturbations and significantly lowers the distance required for successful attacks while maintaining ASR, increasing stealthiness. These findings provide a new direction for exploring the robustness of classifier models.

\subsubsection{Advantages of AS Regularization in Adversarial Attack Methods}
\begin{figure}[H]
    \centering
    \includegraphics[width=1\textwidth]{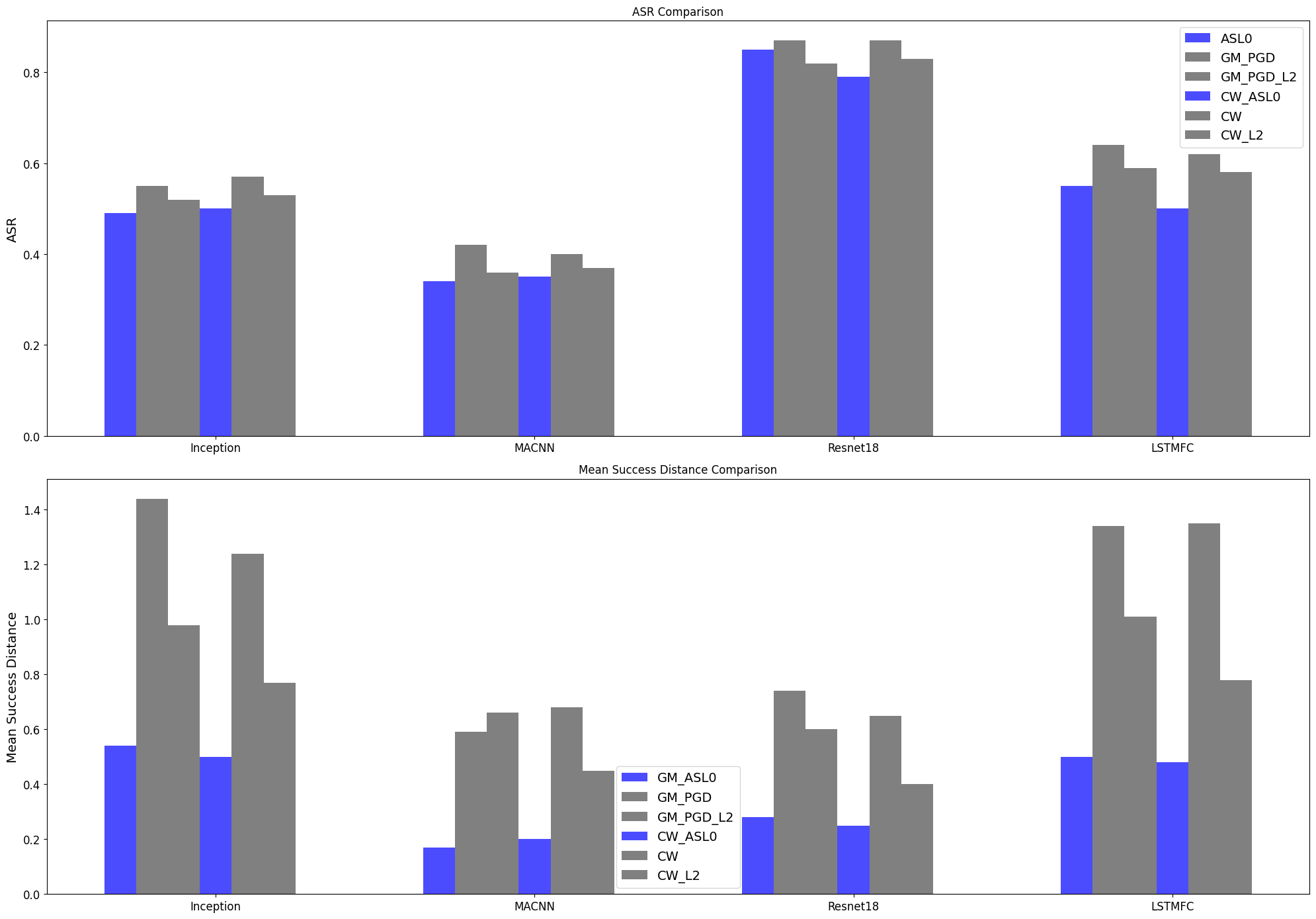}
    \caption{ASR and Success Distance comparison between GM\_PGD, GM\_PGD\_L2, and AS\_GM\_PGD methods across different models.}
    \label{fig:as_gmpgd_advantages}
\end{figure}

Applying AS regularization to GM\_PGD and CW methods demonstrates that ASL0 can significantly reduce the distance required for successful attacks while maintaining the Attack Success Rate (ASR). As depicted in Figure~\ref{fig:as_gmpgd_advantages}, AS regularization maintains competitive ASR across various models and reduces the success distance, indicating a more precise and stealthy adversarial attack.

The experimental results reveal that integrating AS regularisation with the GM\_PGD and CW methods significantly reduces the mean success distance and the overall mean distance while maintaining a relatively stable attack success rate (ASR). For instance, in the Inception model, the mean success distance of the AS\_GM\_PGD method decreased from 1.24 to 0.54, indicating the effectiveness of AS regularisation in enhancing attack precision and stealth. Similar improvements were also observed in the MACNN, Resnet18, and LSTMFC models.

Furthermore, a significant increase in the "Close to 0" metric across all models demonstrates that the AS method can effectively generate sparse and hard-to-detect perturbations, thus enhancing the stealthiness of the attack.

\section{Conclusions and Future Work}

In this study, we have demonstrated an effective adaptive adversarial attack method for time series classification (TSC) models, with a focus on the application of Adaptive Sparse (AS) regularisation techniques. This approach not only successfully showed the capability to maintain high attack success rates while reducing perturbations, but also highlighted the limitations of current defence mechanisms against highly adaptive attacks. This underscores the need for future security strategies to consider the adaptiveness and intelligence of adversarial behaviours.

Our findings prompt a reevaluation of the static nature of existing defence strategies, emphasizing the necessity of developing defence mechanisms that can adapt and respond quickly to the evolution of adversarial tactics. Future work will focus on exploring new regularisation techniques, self-learning and meta-learning algorithms, and other machine-learning paradigms to enhance model robustness and resilience against various attacks. Additionally, the application of AS regularization techniques across different attack methods and model architectures will be investigated to deepen our understanding of their versatility and limitations, and to explore integrating these techniques into TSC models to improve model performance.

\newpage
\section*{Appendix: Model Performance Tables}
The tables below summarize the performance metrics of different deep learning models under adversarial attacks, comparing various regularization strategies.
\begin{longtable}{|c|l|l|l|l|l|}
\caption{Summary of Deep Learning Model Performance under Adversarial Attacks with Different Regularization Strategies.}\label{tbl:model_performance} \\
\hline

\textbf{Model} & \textbf{Metric} & \textbf{ASL0} & \textbf{L1} & \textbf{L2} & \textbf{No Reg.} \\
\hline
\endfirsthead
\multicolumn{6}{c}%
{\tablename\ \thetable\ -- \textit{Continued from previous page}} \\
\hline
\textbf{Model} & \textbf{Metric} & \textbf{ASL0} & \textbf{L1} & \textbf{L2} & \textbf{No Reg.} \\
\hline
\endhead
\hline \multicolumn{6}{r}{\textit{Continued on next page}} \\
\endfoot
\hline
\endlastfoot
\multirow{6}{*}{INCEPTION} & ASR & \textbf{0.59} & 0.60 & 0.60 & 0.60 \\
 & Mean Success Distance & \textbf{0.57} & 0.65 & 0.67 & 0.67 \\
 & Mean Failure Distance & \textbf{1.85} & 2.43 & 2.90 & 2.89 \\
 & Overall Mean Distance & \textbf{0.91} & 1.24 & 1.64 & 1.64 \\
 & duration & \textbf{129.28} & 135.55 & 139.98 & 102.25 \\
 & close to 0 & \textbf{14.48} & 10.62 & 0.21 & 0.22 \\
\hline
\multirow{6}{*}{ResNet18} & ASR & \textbf{0.90} & 0.90 & 0.91 & 0.91 \\
 & Mean Success Distance & \textbf{0.22} & 0.25 & 0.27 & 0.27 \\
 & Mean Failure Distance & \textbf{1.49} & 2.00 & 2.08 & 2.08 \\
 & Overall Mean Distance & \textbf{0.28} & 0.32 & 0.33 & 0.33 \\
 & duration & \textbf{123.83} & 136.61 & 136.91 & 32.26 \\
 & close to 0 & \textbf{27.20} & 37.21 & 0.52 & 0.52 \\
\hline
\multirow{6}{*}{MACNN} & ASR & \textbf{0.40} & 0.38 & 0.57 & 0.58 \\
 & Mean Success Distance & \textbf{0.16} & 0.16 & 0.43 & 0.45 \\
 & Mean Failure Distance & \textbf{0.11} & 0.03 & 0.34 & 0.38 \\
 & Overall Mean Distance & \textbf{0.12} & 0.07 & 0.36 & 0.39 \\
 & duration & \textbf{67.65} & 76.69 & 76.91 & 66.45 \\
 & close to 0 & \textbf{220.70} & 241.59 & 1.78 & 1.31 \\
\hline
\multirow{7}{*}{LSTMFCN} & ASR & \textbf{0.68} & 0.73 & 0.73 & 0.73 \\
 & Mean Success Distance & \textbf{0.44} & 0.56 & 0.62 & 0.62 \\
 & Mean Failure Distance & \textbf{1.20} & 1.90 & 2.06 & 2.06 \\
 & Overall Mean Distance & \textbf{0.65} & 0.84 & 0.89 & 0.89 \\
 & Overall Mean Distance & \textbf{0.12} & 0.07 & 0.36 & 0.39 \\
 & duration & \textbf{39.55} & 42.21 & 43.73 & 30.09 \\
 & close to 0 & \textbf{32.45} & 45.67 & 0.86 & 0.80 \\
\end{longtable}

\end{document}